\documentclass[10pt,twocolumn,letterpaper]{article}

\usepackage{cvpr}
\usepackage{times}
\usepackage{epsfig}
\usepackage{epstopdf}
\usepackage[table]{xcolor}
\usepackage{paralist}
\usepackage{graphics}
\usepackage{epsf}
\usepackage{amsthm, multirow, paralist}
\usepackage{amsmath,amssymb,amsopn}
\usepackage{bm} 
\usepackage{fullpage}
\usepackage{url,color}
\usepackage{graphicx}
\usepackage{xcolor,colortbl}
\usepackage{multirow}
\usepackage{wrapfig}
\usepackage{sidecap}

\usepackage{algorithmicx}
\usepackage{algorithm}
\usepackage{algpseudocode}
\algnewcommand\algorithmicinput{\textbf{Input:}}
\algnewcommand\Input{\item[\algorithmicinput]}
\algnewcommand\algorithmicoutput{\textbf{Output:}}
\algnewcommand\Output{\item[\algorithmicoutput]}

\usepackage[pagebackref=true,breaklinks=true,letterpaper=true,colorlinks,bookmarks=false]{hyperref}

\newcommand{\inner}[2]{\left\langle#1, #2\right\rangle}
\newcommand{\pheadA}[1] {\noindent\textbf{#1.}~} 
\newcommand{\pheadB}[1] {\vspace{1mm}\noindent\textbf{#1.}~} 

\def\cvprPaperID{950} 

\cvprfinalcopy 

\def\cvprPaperID{****} 

\begin{document}

\title{Learning Attributes Equals \\ Multi-Source Domain Generalization}

\author{Chuang Gan\\
IIIS, Tsinghua University\\
{\tt\small ganchuang1990@gmail.com}
\and
Tianbao Yang\\
University of Iowa\\
{\tt\small tianbao-yang@uiowa.edu}
\and
Boqing Gong\\
CRCV, U.\ of Central Florida\\
{\tt\small bgong@crcv.ucf.edu}
}

\maketitle
 \thispagestyle{empty}

\begin{abstract}

    Attributes possess appealing properties and benefit many  computer vision problems, such as object recognition, learning with humans in the loop, and  image retrieval. Whereas the existing work mainly pursues utilizing attributes for various computer vision problems, we contend that the most basic problem---how to accurately and robustly detect attributes from images---has been left under explored.  Especially, the existing work rarely explicitly tackles the need that attribute detectors should generalize well across different categories, including those previously unseen.
    Noting that this is analogous to the objective of multi-source domain generalization, if we treat each category as a domain, we provide a novel perspective to attribute detection and propose to gear the techniques in multi-source domain generalization  for the purpose of learning cross-category generalizable attribute detectors. We validate our understanding and approach with extensive experiments on four challenging datasets and three different problems.



\end{abstract}

\section{Introduction}
\label{sec:1}

Visual attributes are middle-level concepts which humans use to describe objects, human faces, scenes, activities, and so on (e.g., four-legged, smiley, outdoor, and crowded). A major appeal of attributes is that they are not only human-nameable but also machine-detectable, making it possible to serve as the building blocks to describe instances~\cite{farhadi2009describing, kumar2009attribute, patterson2012sun,parikh2011relative}, teach machines to recognize previously unseen classes by zero-shot learning~\cite{lampert2009learning,palatucci2009zero}, or offer a natural human-computer interactions channel for image/video search~\cite{siddiquie2011image,yu2012weak,kovashka2012whittlesearch,sun2015automatic}.

\begin{figure}[t]
   \centering
   \includegraphics[width = 1.0\linewidth]{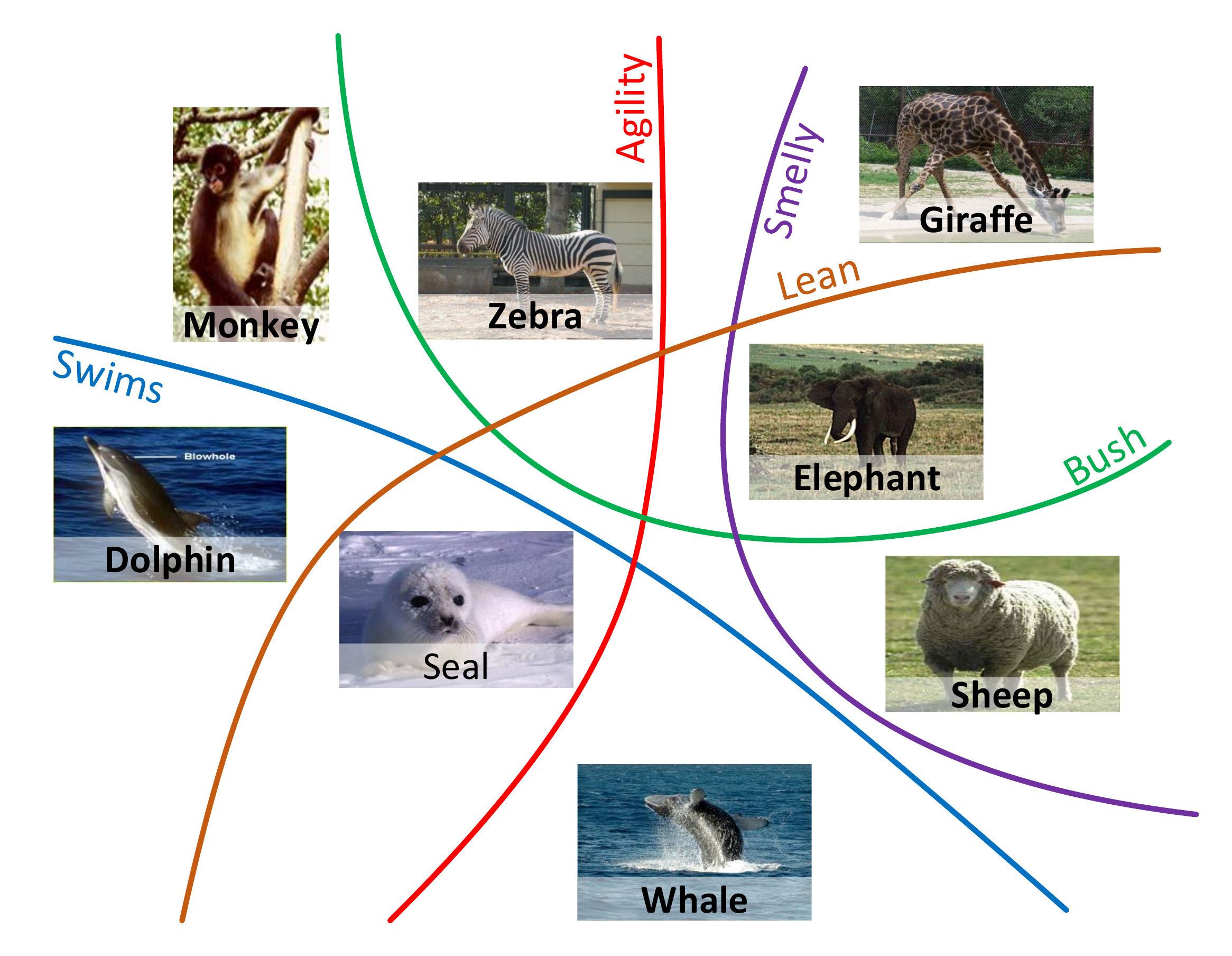}
    \caption{The boundaries between middle-level attributes and high-level object classes cross each other. We thus do not expect that
the features originally learned for separating elephant, sheep, and giraffe could also be optimal for detecting
the attribute ``bush'', which is shared by them. We propose to understand attribute detection as multi-source domain generalization and to explicitly break the class boundaries in order to learn high-quality attribute detectors. }
   \label{fig:highlight}
\end{figure}

However, we contend that the long-standing pursuit after utilizing attributes for various computer vision problems \textbf{has left the most basic problem---how to accurately and robustly detect attributes from images or videos---far from being solved.} Especially, the existing work rarely explicitly tackles the need that \textbf{attribute detectors should generalize well across different categories, including those previously unseen ones.} For instance, the attribute detector ``four-legged" is expected to correctly tell a giant panda is four-legged even if it is trained from the images of horses, cows, zebras, and pigs (i.e., no pandas).

Indeed, most of the existing \emph{attribute} detectors~\cite{lampert2009learning,farhadi2009describing, kumar2009attribute,chen2012describing,hwang2011sharing,jayaraman2014decorrelating,chen2014inferring,chen2014predicting,choi2014scene,vedaldi2014understanding,huang2015learning} are built using features engineered or learned for \emph{object} recognition together with off-shelf machine learning classifiers---without tailoring them to reflect the idiosyncrasies of attributes. This is suboptimal; the successful techniques on object recognition do not necessarily apply to attributes learning mainly for two reasons. First, attributes are in a different semantic space as opposed to objects; they are in the \emph{middle} of low-level visual cues and the high-level object labels. Second, attribute detection can even be considered as an \emph{orthogonal} task to object recognition, in that attributes are shared by different objects (e.g., zebras, lions, and mice are all ``furry'') and distinctive attributes are present in the same object (e.g., a car is boxy and has wheels). As shown in Figure~\ref{fig:highlight}, the boundaries between attributes and between object categories cross each other. Therefore, we do not expect that the features originally learned for separating elephant, sheep, and giraffe could also be optimal for detecting the attribute ``bush'', which is shared by them. 

In this paper, we propose to re-examine the fundamental attribute detection problem and aim to develop an attribute-oriented feature representation, such that one can conveniently apply off-shelf classifiers to obtain high-quality attribute detectors. We expect that the detectors learned from our new representation are capable of breaking the boundaries of object categories and generalizing well across both seen and unseen classes. To this end,  we cast \textbf{attribute detection as a multi-source domain generalization problem}~\cite{muandet2013domain,xu2014exploiting,niu2015visual} by noting that the desired properties from attributes are analogous to the objective of the latter.

Particularly, a domain refers to an underlying data distribution. Multi-source domain generalization aims to extract knowledge from several \emph{related} source domains such that it is applicable to different domains, especially to those unseen at the training stage. This is in accordance with our objective for learning cross-category generalizable attributes detectors, if we consider each category as a distinctive domain.

Motivated by this observation, we employ the Unsupervised Domain-Invariant Component Analysis (UDICA)~\cite{muandet2013domain} as the basic building block for our approach. The key principle of UDICA is that minimizing the distributional variance of different domains---categories in our context, can improve the cross-domain (cross-category) generalization capabilities of the classifiers. A supervised extension to UDICA was introduced in~\cite{muandet2013domain} depending on the inverse of a covariance operator as well as some mild assumptions. However, the inverse operation is both computationally expensive and unstable in practice. We instead propose to use the alternative of centered kernel alignment~\cite{cortes2012algorithms} to account for the attribute labeling information. We show that the centered kernel alignment can be seamlessly integrated with UDICA, enabling us to learn both category-invariant and attribute-discriminative feature representations.


Our approach takes as input the features of the training images, their class (domain) labels, as well as their attribute labels. It operates upon kernels derived from the input data and learns a kernel projection to ``distill''  category-invariant and attribute-discriminative signals embedded in the original features. The overall output is a new feature vector for each image, which can be readily used in traditional machine learning models like SVMs for training the cross-category generalizable attribute detectors.

The contributions of the paper are summarized below.

\begin{compactitem}
	\item To the best of our knowledge, this work is the first attempt to tackle attribute detection from the multi-source domain generalization point of view. This enables us to explicitly model the need  that the attribute detectors should transcend different  categories and generalize to previously unseen ones.


    \item We introduce the centered kernel alignment to UDICA and arrive at an integrated method to  strengthen the discriminative power of the learned attributes on one hand, and eliminate the domain differences between categories on the other hand.


    \item We test our approach to four datasets: Animal With Attributes~\cite{lampert2009learning}, Caltech-UCSD Birds~\cite{wah2011caltech}, aPascal-aYahoo~\cite{farhadi2009describing}, and UCF101~\cite{ucf101}, and test the learned representations on three tasks: attribute detection itself, zero-shot learning, and image retrieval. Our results are significantly better than those of competitive baselines, verifying the effectiveness of the new perspective for solving attribute detection as domain generalization. \vspace{0.05in}


\end{compactitem}

The rest of this paper is organized as follows. In Section~\ref{sec:2}, we review related work in attribute detection, domain generalization, and domain adaptation. Section~\ref{sec:3} and section~\ref{sec:4} present the attribute learning framework. The experimental settings and evaluation results are presented in Section~\ref{sec:5}. Section~\ref{sec:6} concludes the paper.

\section{Related work and background}
\label{sec:2}
Our approach is related to two separate research areas, attribute detection and domain adaptation/generalization. We unify them in this work.


\paragraph{Attributes learning.}
Earlier work on attribute detection mainly focused on modeling the correlations among attributes~\cite{chen2012describing,hwang2011sharing,jayaraman2014decorrelating,chen2014inferring,chen2014predicting,choi2014scene,vedaldi2014understanding,huang2015learning}, localizing some special part-related attributes (e.g., tails of mammals)~\cite{berg2010automatic,bourdev2011describing,joo2013human,berg2013poof,zhang2014panda,sandeep2014relative,duan2012discovering}, and the relationship between attributes and categories~\cite{wang2010discriminative,mahajan2011joint,hwang2011sharing,parikh2011interactively}. Some recent work has applied deep models to  attribute detection~\cite{chen2015deep,zhang2014panda,luo2013deep,escorcia2015relationship}. None of these methods explicitly model the cross-category generalization of the attributes, except the one by Farhadi et al.~\cite{farhadi2009describing} where the authors select features within each category to down-weight category-specific cues. Likely due to the fact that the attribute and category cues are interplayed, their feature selection procedure only gives limited gain. We propose to overcome this challenge by investigating all categories together and employing nonlinear mapping functions.

Attributes possess versatile properties and benefit a wide range of challenging computer vision tasks.  They serve as the basic building blocks for one to compose  categories (e.g., different objects)~\cite{lampert2009learning,palatucci2009zero,yu2010attribute,farhadi2010attribute,kankuekul2012online,wang2013unified,akata2013label,jayaraman2014zero} and describe instances~\cite{farhadi2009describing, kumar2009attribute, patterson2012sun,parikh2011relative,wang2013motionlets}, enabling knowledge transfer between them. Attributes also reveal the rich structures underlying categories and are thus often employed to regulate machine learning models for visual recognition~\cite{su2010improving,torresani2010efficient,li2010object,shrivastava2012constrained,hwang2014unified,gan2016recognizing,gan2015exploring,shi2014weakly}.
Moreover, attributes offer a natural human-computer interaction channel for visual recognition with humans in the loop~\cite{branson2010visual}, relevance feedback in image retrieval~\cite{kumar2009attribute,siddiquie2011image,parikh2011relative,scheirer2012multi,yu2012weak,kovashka2012whittlesearch,rastegari2013multi,gong2013learning,tao2015attributes},
and active learning~\cite{kovashka2011actively,parkash2012attributes,biswas2013simultaneous,liang2014beyond,lad2014interactively}.
In this paper, we test the proposed approach on both attribute detection and its applications to zero-shot learning and image retrieval.

\paragraph{Domain generalization and adaptation.}


Domain generalization is still at its early developing stage.  A feature
projection-based algorithm, Domain-Invariant Component Analysis
(DICA), was  introduced in~\cite{muandet2013domain} to learn by minimizing the variance of the source domains.
Recently, domain generation has been introduce into computer vision community for object recognition~\cite{xu2014exploiting,ghifary2015domain} and video recognition~\cite{niu2015visual}. We propose to gear multi-source domain generalization techniques for the purpose of learning cross-category generalizable attribute detectors. Multi-source domain adaptation~\cite{mansour2009domain,hoffman2012discovering,gong2013reshaping,sun2011two,duan2009domain} is related to our approach if we consider a transductive setting (i.e., the learner has access to the test data). While it assumes a single target domain, in attribute detection the test data are often sampled from more than one unseen domain.

\subsection{Background on distributional variance} \label{sDV}
Denote by $\mathcal{H}$ and $k(\cdot,\cdot)$ respectively a Reproducing Kernel Hilbert Space and its associated kernel function. For an arbitrary distribution $P_y(\bm{x})$ indexed by $y\in\mathcal{Y}$, the following mapping,
\begin{align}
\mu[P_y(\bm{x})] = \int k(\bm{x},\cdot) \text{d}P_y(\bm{x}) \triangleq \mu_y
\end{align}
is injective if $k$ is a characteristic kernel~\cite{smola2007hilbert,gretton2006kernel,sriperumbudur2010hilbert}. In other words, the kernel mean map $\mu_y$ in the RKHS $\mathcal{H}$ preserves all the statistical information of $P_y(\bm{x})$.

The distributional variance follows naturally,
\begin{align}
\mathbb{V}(\mathcal{Y}) = \frac{1}{|\mathcal{Y}|} \sum_{y\in\mathcal{Y}} \|\mu_y - \mu_0\|_{\mathcal{H}}^2,  \; \widehat{\mathbb{V}}(\mathcal{Y})=\text{tr}(KQ),
\end{align}
where $\mu_0$ is the map of the mean of all the distributions in $\mathcal{Y}$. In practice, we do not have access to the distributions. Instead, we observe the samples $S_y, y\in\mathcal{Y}$ each drawn from a distribution $P_y(\bm{x})$ and can thus empirically estimate the distributional variance by $\widehat{\mathbb{V}}(\mathcal{Y})=\text{tr}(KQ)$. Here $K$ is the (centered)\footnote{All kernels discussed in this paper have been centered~\cite{cortes2012algorithms}.} kernel matrix over all the samples, and $Q$ collects the coefficients which depend on only the numbers of samples.  We refer the readers to~\cite{muandet2013domain} for more details including the consistency between the distributional variance $\mathbb{V}$ and its estimate $\widehat{\mathbb{V}}$.


\section{Attribute detection}
\label{sec:3}
This section formalizes attribute detection and shows its in-depth connection to domain generalization.

\paragraph{Problem statement.} Suppose that we have access to an annotated dataset of $\mathsf{M}$ images. They are in the form of $(\bm{x}_m,\bm{a}_m, y_m)$ where $\bm{x}_m\in\mathbb{R}^\mathsf{D}$ is the feature vector extracted from the $m$-th image $I_m$, $m\in[\mathsf{M}]\triangleq\{1,2,\cdots,\mathsf{M}\}$. Two types of annotations are provided for each image, the category label $y_m\in[{\mathsf{C}}]$ and the attribute annotations $\bm{a}_m\in\{0,1\}^\mathsf{A}$. Though we use binary attributes (e.g., the presence or absence of stripes) to in this paper for clarity, it is straightforward to extend our  approach to multi-way and continuous-valued attributes. Note that a particular attribute $\bm{a}_{m}^i$ could appear in many categories (e.g., zebras, cows, giant pandas, lions, and mice are all furry). Moreover, there may be test images from previously unseen categories $\{\mathsf{C}+1,\mathsf{C}+2,\cdots\}$ for example in zero-shot learning. Our objective is to learn accurate and robust attribute detectors $\mathcal{C}(\bm{x}_m)\in\{0,1\}^\mathsf{A}$ to well generalize across different categories, especially to be able to perform well on the unseen classes.

\begin{figure}
  \centering
  \includegraphics[width=0.48\textwidth]{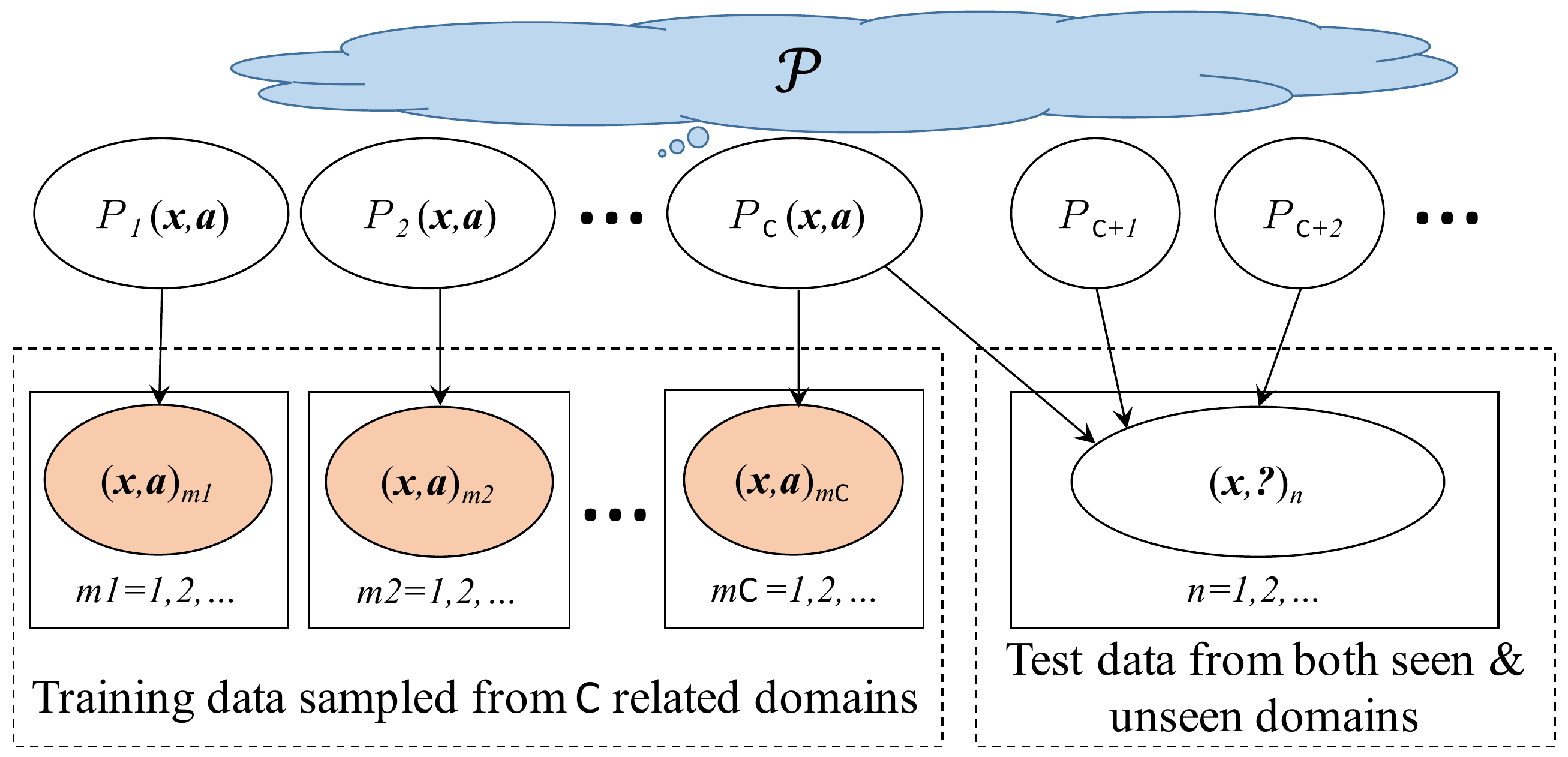}
  \caption{Attribute detection as multi-source domain generalization. Given  labeled data sampled from several categories/domains, i.e., distributions $P_y(\bm{x},\bm{a}), y\in[{\mathsf{C}}]$ over image representations $\bm{x}$ and attribute labels $\bm{a}$, we extract knowledge useful for attribute detection and applicable to different domains/categories, especially to previously unseen ones $P_{{\mathsf{C}}+1},P_{{\mathsf{C}}+2},\cdots$. The domains are assumed related and sampled from a common distribution $\mathcal{P}$.} \label{fDG}
\end{figure}

\paragraph{Attribute detection as domain generalization\\---A new perspective.}
In this paper, we understand attribute detection as a domain generalization problem. A domain refers to an underlying data distribution. In our context, it refers to the distribution $P_y(\bm{x},\bm{a})$ of a category $y\in[\mathsf{C}]$ over the input $\bm{x}$ and attribute labels $\bm{a}$. As shown in Figure~\ref{fDG}, the domains/categories are assumed to be related and are sampled from a common distribution $\mathcal{P}$. This is reasonable considering that images and categories can often be organized in a hierarchy. Thanks to the relationships between different categories, we expect to learn new image representations for attribute detection, such that the corresponding detectors will perform well on both seen and unseen classes.

\section{Approach}
\label{sec:4}
Our key idea is to find a feature transformation of the input $\bm{x}$ to eliminate the mismatches between different domains/categories in terms of their marginal distributions over the input, whereas ideally we should consider the joint distributions $P_y(\bm{x},\bm{a})$, $y\in[\mathsf{C}]$. In particular, we use Unsupervised Domain Invariant Component Analysis (UDICA)~\cite{muandet2013domain} and centered kernel alignment~\cite{cortes2012algorithms} for this purpose. Note that modeling the marginal distributions $P_y(\bm{x})$ is a common practice in domain generalization~\cite{muandet2013domain,xu2014exploiting,ghifary2015domain} and domain adaptation~\cite{huang2006correcting,pan2011domain,gong2012geodesic} and performs well in many applications. We leave investigating the joint distributions $P_y(\bm{x},\bm{a})$ for future work.

Next, we present how to integrate UDICA and centered kernel alignment. Jointly they give rise to new feature representations which account for both attribute discriminativeness and cross-category generalizability.


\subsection{UDICA} \label{sUDICA}

The projection from one RKHS to another results in the following transformation of the kernel matrices, $\mathbb{R}^{\mathsf{M}\times\mathsf{M}}\ni K\mapsto \widetilde{K}=KBB^TK\in\mathbb{R}^{\mathsf{M}\times\mathsf{M}}$~\cite{scholkopf1997kernel}. As a result, one can take $(KB)$ as the empirical kernel map, i.e., consider the $m$-th row of $(KB)$ as the new feature representations of image $I_m$ and plug them into any linear classifiers.
UDICA learns the transformation $B$ by imposing the following properties.

\paragraph{Minimizing distributional variance.} The empirical distributional variance  (cf.\ Section~\ref{sDV})  between different domains/categories becomes the following in our context,
\begin{align}
\mathbb{V}_B([\mathsf{C}]) = \text{tr}(\widetilde{K}Q) = \text{tr}(B^TKQKB).
\end{align}
Intuitively, the domains would be perfectly matched when the variance is 0. Since there are many seen categories, each as a domain, we expect the learned projection to be generalizable and work well for the unseen classes as well.

\paragraph{Maximizing data variance.} Starting from the empirical kernel map $(KB)$, it is not difficult to see that the data covariance is $(KB)^T(KB)/\mathsf{M}$ and the variance is
\begin{align}
\mathbb{V}_B([\mathsf{M}])= \text{tr}(B^TK^2B) / \mathsf{M}.
\end{align}

\paragraph{Regularizing the transformation.} UDICA regularizes the transformation by minimizing
\begin{align}
\mathcal{R}(B) = \text{tr}(B^TKB).
\end{align}
Alternatively, one can use the Frobenius norm $\|B\|_F$, as did in~\cite{pan2011domain}, to constrain the complexity of $B$.

Combining the above criteria, we arrive at the following problem,
\begin{align}
    \max_{B}  \quad  \frac {\text{tr}(B^{T}K^{2}B) / \mathsf{M}} {\text{tr} (B^{T}KQKB+B^{T}KB)}, \label{eDICA}
\end{align}
where the nominator corresponds to the data variance and the denominator sums up the distributional variance and the regularization over $B$.

By solving the above problem, we are essentially blurring the boundaries between different categories and match the classes with each other, due to the distributional variance term in the denominator. This thus eliminates the barrier for attribute detectors to generalize in various classes. Our experiments verify the effectiveness the learned new representations $(KB)$. Nonetheless, we can further improve the performance by modeling the attribute labels using centered kernel alignment.

\subsection{Centered kernel alignment}
Note that our training data are in the form of $(\bm{x}_m,\bm{a}_m, y_m), m\in[\mathsf{M}]$. For each image there are multiple attribute labels which may be highly correlated. Besides, we would like to stick to kernel methods to be consistent with our choice of UDICA---indeed, the distributional variance is best implemented by kernel methods (cf.\ Section~\ref{sDV}). These considerations lead to our decision on using kernel alignment~\cite{cortes2012algorithms} to model the multi-attribute supervised information.

Let $L_{m,m'}=\inner{\bm{a}_m}{\bm{a}_{m'}}$ be the kernel matrix over the attributes. Since $L$ is computed directly from the attribute labels, it preserves the correlations among them and serves as the ``perfect'' target kernel for the transformed kernel $\widetilde{K}=KBB^TK$ to align to. The centered kernel alignment is then computed by,
\begin{align}
\rho(\widetilde{K}, L) =  \frac{\text{tr}(\widetilde{K} L)}{\sqrt{\text{tr}(\widetilde{K} \widetilde{K})} \sqrt{\text{tr}(L L)}} \label{eKernelAlign}
\end{align}
where we abuse the notation $L$ slightly to denote that it is centered~\cite{cortes2012algorithms}.

We would like to integrate the kernel alignment with UDICA in a unified optimization problem. To this end, firstly it is safe to drop $\text{tr}(L L)$ in eq.~(\ref{eKernelAlign}) since it has nothing to do with the projection $B$ we are learning. Moreover, note that the role of ${\text{tr}(\widetilde{K} \widetilde{K})}$ duplicates with the regularization in eq.~(\ref{eDICA}) to some extent, as it is mainly to avoid trivial solutions for the kernel alignment.  We thus only add $\text{tr}(\widetilde{K} L)$ to the nominator of UDICA,
\begin{align}
    \max_{B}  \quad  \frac {\text{tr}(\gamma B^{T}K^{2}B / \mathsf{M} + (1-\gamma)B^TKLKB)} {\text{tr} (B^{T}KQKB+B^{T}KB)}, \label{eAll}
\end{align}
where $\gamma\in[0,1]$ balances the data variance and the kernel alignment with the supervised attribute labeling information. We cross-validate $\gamma$  in our experiments. We name this formulation {\bf KDICA}, which couples the centered kernel alignment and UDICA. The former closely tracks the attribute discriminative information and the latter facilitates the cross-category generalization of the attribute detectors to be trained upon KDICA.

\paragraph{Optimization.} By writing out the Lagrangian of the formalized problem (eq.~(\ref{eAll})) and then setting the derivative with respect to $B$ to 0, we arrive at a generalized eigen-decomposition problem,
\begin{align}
& \;\left(\gamma K^{2} / \mathsf{M} + (1-\gamma)KLK\right)B  \notag\\
 = & \;\left(KQK+K\right)B\Gamma,
\label{eq:9}
\end{align}
where $\Gamma$ is a diagonal matrix containing all the eigenvalues (Lagrangian multipliers). We find the solution $B$ as the Leading eigen-vectors. The number of eigen-vectors is cross-validated in our experiments. Again, we remind that $(KB)$ serves as the new feature representations of the images for training attribute detectors. The details of our proposed framework has been shown in algorithm~\ref{algo:KDICA}.

\begin{algorithm}[t]
\caption{Kernel-alignment Domain-Invariant Component Analysis (KDICA). }
\label{algo:KDICA}
\begin{algorithmic}[1]
	\Input Parameters $\gamma$ and $\mathsf{b} \ll \mathsf{M}$. Training data $\mathcal{S} = \{(\bm{x}_m, y_m, \bm{a}_m)\}_{m=1}^\mathsf{M}$
	\Output Projection $B_{ \mathsf{M} \times \mathsf{b}}$
	\State Calculate gram matrix $[K_{ij}] = k(\bm{x}_i, \bm{x}_j)$ and $[L_{ij}] = l(\bm{a}_i, \bm{a}_j)$
	\State Solve: \newline
      $ (\gamma K^{2} / \mathsf{M} + (1-\gamma)KLK)B = (KQK+K)B\Gamma$.
 	\State Output $B$ and $\widetilde{K} \leftarrow KBB^{\mathrm{T}} K$
 	\State Use $(KB)$ as if they are the features to learn linear classifiers and $\widetilde{K}$ for kernelized classifiers
\end{algorithmic}
\end{algorithm}

\section{Experiment}
\label{sec:5}
This section presents our experimental results on  four benchmark datasets. We test our approach for both the immediate task of attribute detection and two other problems, zero-shot learning and image retrieval, which could benefit from high-quality attribute detectors.

\begin{table*}
    \centering

    \begin{tabular}{c|cccc}
    \hline
        Approcahes & AWA & CUB & a-Yahoo & UCF101 \\
    \hline

        IAP~\cite{lampert2009learning} & $\mathrm{74.0/{79.2}^{*}}$ &  $\mathrm{{74.9}^{*}}$ &  -- & --       \\

        ALE~\cite{akata2013label} & 65.7 &  60.3 & --  & -- \\

        HAP~\cite{choi2014scene} &   $\mathrm{74.0/{79.1}^{*}}$    &     $\mathrm{68.5/{74.1}^{*}}$  & $\mathrm{{58.2}^{*}}$ & 72.1  $\pm$ 1.1 \\

       $\mathrm{CSHAP}_G$~\cite{choi2014scene} & $\mathrm{74.3/{79.4}^{*}}$  & $\mathrm{62.7/{74.6}^{*}}$  & $\mathrm{{58.2}^{*}}$&  72.3 $\pm$ 1.0\\

       $\mathrm{CSHAP}_H$~\cite{choi2014scene}  & $\mathrm{74.0/{79.0}^{*}}$  & $\mathrm{68.5/{73.4}^{*}}$ & $\mathrm{{65.2}^{*}}$ &72.4  $\pm$ 1.1 \\

         DAP~\cite{lampert2009learning} & $\mathrm{72.8/{78.9}^{*}}$      &  $\mathrm{61.8/{72.1 }^{*}}$ & $\mathrm{{77.4}^{*}}$ &  71.8  $\pm$ 1.2 \\

    \hline
        UDICA (Ours)    & \textbf{83.9}    & \textbf{76.0}  & \textbf{82.3}   & \textbf{74.3 $\pm$ 1.3} \\

        KDICA (Ours) & \textbf{84.4} &   \textbf{76.4}    & \textbf{84.7}   &  \textbf{75.5 $\pm$ 1.1} \\
    \hline

\hline
\end{tabular}
    \vspace{1mm}
 \caption{Average Attribute Prediction Accuracy (\%, in AUC.)}
\label{tab:attribute}
\end{table*}

\subsection{Experiment setup}
\label{sec:4.1}
\pheadA{Dataset}
We use four datasets to validate the proposed approach; three of them contain images for object and scene recognition and the last one contains videos for action recognition. (a)~The \textbf{Animal with attribute (AWA)}~\cite{lampert2009learning} dataset comprises of 30,475 images belonging to 50 animal classes. Each class is annotated with 85 attributes. Following the standard split by the dataset, we divide the dataset into 40 classes (24,295 images) to be used for training and 10 classes (6,180 images)  for testing.  (b)~\textbf{Caltech-UCSD Birds 2011 (CUB)}~\cite{wah2011caltech} is a dataset with fine-grained objects. There are 11,788 images of 200 different bird classes in CUB. Each class is annotated with 312 binary attributes. We split the dataset as suggested in~\cite{akata2013label} to facilitate direct comparison (150 classes for training and 50 classes for testing). (c)~\textbf{aPascal-aYahoo}~\cite{farhadi2009describing} consists of two attribute datases: the a-PASCAL dataset, which contains 12,695 images (6,340 for training and 6,355 for testing) collected for the Pascal VOC 2008 challenge, and a-Yahoo including 2,644 test images. Each images is annotated with 64 attributes. There are 20 object classes in a-Pascal and 12 in a-Yahoo and they are disjoint. Following the settings of~\cite{siddiquie2011image,yu2012weak}, we use the pre-defined training images in a-Pascal  as the training set and test on a-Yahoo. (d)~\textbf{UCF101 dataset}~\cite{ucf101} is a large dateset for video action recognition. It contains 13,320 videos of 101 action classes. Each action class comes with 115 attributes. The videos are collected from YouTube with large variations in camera motion, object appearance, viewpoint, cluttered background, and illumination conditions.  We run 10 rounds of experiments each with a random split of 81/20 classes  for the training/testing sets, and then report the averaged results.

\pheadA{Features}
For the first three image datasets, we use the Convolutional Neural Network (CNN) implementation provided by Caffe~\cite{caffe}, particularly with the 19-layer network architecture and parameters from Oxford~\cite{Verydeep}, to extract  4,096-dimensional CNN feature representations from images ($i.e.$, the activations of the first fully-connected layer fc6). For the UCF101 dataset, we use the 3D CNN (C3D)~\cite{C3D} pre-trained on the Sport 1M dataset~\cite{Sports1M} to construct video-clip features from both spatial and temporal
dimensions. We then use average pooling to obtain the video-level representations. We $\ell_2$ normalize the feature representations in the following experiments.

\pheadA{Implementation details}
We choose the  Gaussian RBF kernel and fix the bandwidth parameter as 1 for our approach to learning new image representations. After that, to train the attribute detectors, we input the learned representations into standard linear Support Vector Machines (see the empirical kernel map in Setcion~\ref{sUDICA}).  There are two free hyper-parameters when we train the detectors using the representations learned through UDICA, the hyper-parameter $C$ in SVM and the number $\mathsf{b}$ of leading eigen-vectors in UDICA. We use five-fold cross-validation to choose the best values for $C$ and $\mathsf{b}$ respectively from $\{0.01, 0.1 , 1, 10, 100\}$ and $\{30, 50, 70, 90, 110, 130, 150\}$. We use the same $C$ and $\mathsf{b}$ for KDICA and only cross-validate  $\gamma$  in equation~(\ref{eq:9}) from $\{0.2, 0.5, 0.8\}$ to learn the SVM based attribute detectors with KDICA.


\begin{figure*}[t]
   \centering
   \includegraphics[width = 1.0\linewidth]{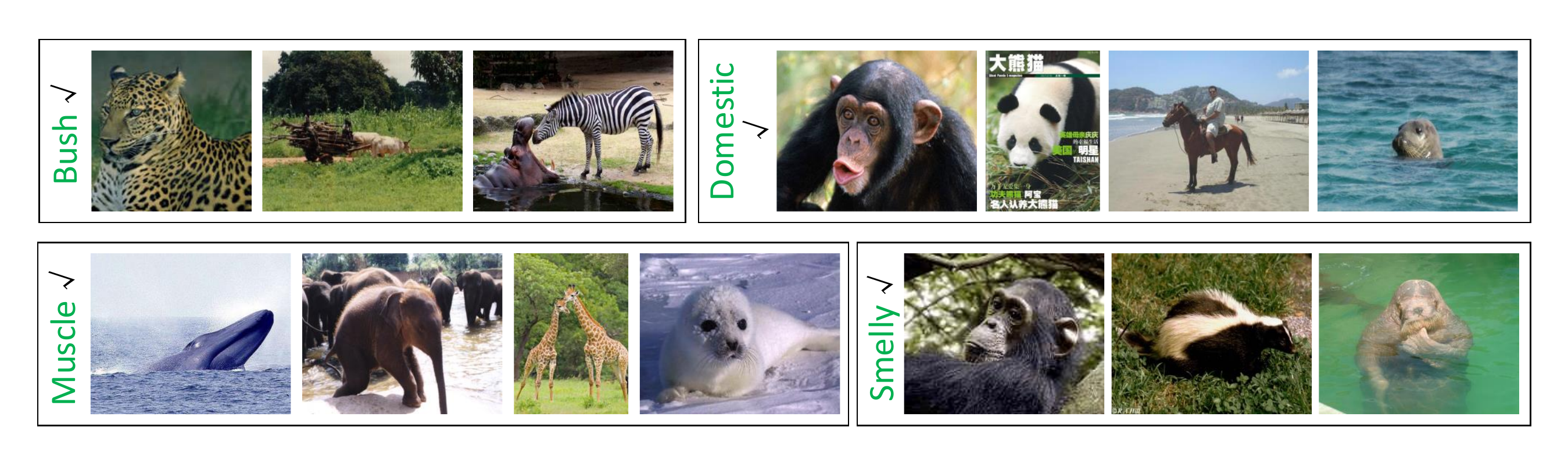}
    \caption{Some attributes on which the proposed KDICA significantly improves the performance of DAP. }
   \label{fig:good}

\end{figure*}

\pheadA{Evaluation}
We first test our approach to attribute detection on all the four datasets (AWA, CUB, aPascal-aYahoo, and UCF101). To see how much the other tasks, which involve attributes, can gain from higher-quality attribute detectors, we further conduct zero-shot learning~\cite{palatucci2009zero,lampert2009learning} experiments on AWA, CUB, and UCF101, and multi-attribute based image retrieval on AWA. We evaluate the results of attribute detection and image retrieval by the averaged Area Under ROC Curve (AUC), the higher the better, and the results of zero-shot learning by classification accuracy.


\subsection{Attribute prediction}
\label{sec:5.2}

Table~\ref{tab:attribute} presents the attribute prediction performance of our approaches and several competitive baselines. In particular, we compare with four state-of-the-art attribute detection methods: Directly Attribute Prediction (DAP)~\cite{lampert2009learning}, Indirectly Attribute Prediction (IAP)~\cite{lampert2009learning}, Attribute Label Embedding (ALE)~\cite{akata2013label}, and Hypergraph-regularized Attribute Predictors (HAP)~\cite{choi2014scene}. Note that we can directly contrast our methods with DAP to see the effectiveness of the learned new representations, because they share the same input and classifiers and only differ in that we learn the new attribute-discriminative and category-invariant feature representations. The IAP model first maps any input to the seen classes and then predicts the attributes on top of those. The ALE method unifies attribute prediction with object class prediction instead of directly optimizing with respect to attributes. We thus do not expect it to perform quite well on the attribute prediction task. HAP explores the correlations among attributes explicitly by hyper-graphs, while we achieve this implicitly in the kernel alignment. Additionally, we also show the results of $\mathrm{CSHAP}_G$ and $\mathrm{CSHAP}_H$, two variations of HAP to model class labels.

We include in Table~\ref{tab:attribute} both the results of these methods reported in the original papers, when they are available, and those we obtained (marked by  `*') by running the source code provided by the authors. We use the same CNN features (for AWA, CUB, and aPascal-aYahoo) and C3D features (for UCF101) we extracted for the baselines and our approach.

\begin{table}
    \centering
    \footnotesize

    \begin{tabular}{c|ccc}
    \hline
        Approcahes & AWA & CUB & UCF101 \\
    \hline

        ALE~\cite{akata2013label} & 37.4 &  18.0 & -- \\

        HLE~\cite{akata2013label} &39.0 & 12.1 & --\\

        AHLE~\cite{akata2013label} & 43.5 & 17.0 & --\\

        DA~\cite{jayaraman2014decorrelating} & 30.6 & -- & --\\

        MLA~\cite{fu2014multimodal}& 41.3 &         --      &--   \\

        ZSRF~\cite{jayaraman2014zero} &48.7 & -- & --  \\

        SM~\cite{fu2015zero} & 66.0 &-- & --   \\

        Embedding~\cite{akata2015evaluation} & 60.1 & 29.9 &-\\

        \hline

       IAP~\cite{lampert2009learning} & $\mathrm{42.2/{49.4}^{*}}$   &  $\mathrm{4.6/{34.9}^{*}}$   &  --         \\

        HAP~\cite{choi2014scene} &   $\mathrm{45.0/{55.6}^{*}}$    &     $\mathrm{17.5/{40.7}^{*}}$  & -- \\

       $\mathrm{CSHAP}_G$~\cite{choi2014scene} & $\mathrm{45.0/{54.5}^{*}}$  & $\mathrm{17.5/{38.7}^{*}}$ & --\\

       $\mathrm{CSHAP}_H$~\cite{choi2014scene}  & $\mathrm{45.6/{53.3}^{*}}$  & $\mathrm{17.5/{36.9}^{*}}$& -- \\

         DAP~\cite{lampert2009learning} & $\mathrm{41.2/{58.9}^{*}}$      &  $\mathrm{10.5/{39.8 }^{*}}$ & 26.8 $\pm$ 1.1\\

    \hline
        UDCIA (Ours)    & \textbf{63.6}    & \textbf{42.4}  & \textbf{29.6 $\pm$ 1.2}    \\

        KDCIA (Ours) & \textbf{73.8} &   \textbf{43.7}    & \textbf{31.1 $\pm$ 0.8}     \\
    \hline

\hline
\end{tabular}
\vspace{3mm}
 \caption{Zero-shot recognition performances. (\%, in accuracy)}
\label{tab:zero}
\end{table}

\pheadB{Overall results}
From  Table~\ref{tab:attribute}, we can find that UDCIA and KDICA outperform all the baselines on all the four datasets. More specifically, the relative accuracy gains of UDCIA over DAP are 6.3\% on the AWA dataset and 5.4\% on the CUB dateset, respectively, under the same feature and experimental settings. These clearly validate our assumption that blurring the category boundaries improves the generalizabilities of attribute detectors to previously unseen categories.  The KDICA with centered kernel alignment  is slightly better than the UDICA approach by incorporating attribute discriminative signals into the new feature representations. Delving into the per-unseen-class attribute detection result, we find that our KDICA-based approach improves the results of DAP for 71 out of 85 attributes on AWA  and 272 out of 312  on CUB.

\begin{figure}[t]
   \centering
   \includegraphics[width = 1.0\linewidth]{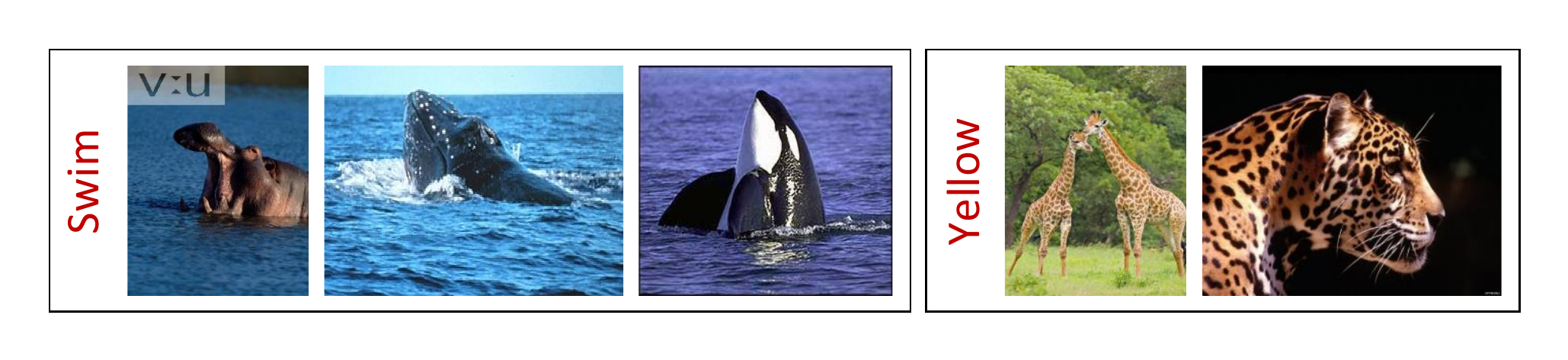}
    \caption{Example attributes that KDICA could not improve the detection accuracy over the traditional DAP approach.}
    \vspace{3mm}
   \label{fig:bad}
\end{figure}

\pheadB{When domain generalization helps}
We give some qualitative analyses using Figure~\ref{fig:good} and~\ref{fig:bad} here. For the attributes in Figure~\ref{fig:good}, the proposed KDICA significantly improves the performance of the DAP approach. Those attributes (``muscle'', ``domestic'', etc.) appear in visually quite different object categories. It seems like breaking the category boundaries is necessary in this case in order to make the attribute detectors generalize to the unseen classes. On the other hand, Figure~\ref{fig:bad} shows the attributes for which our approach can hardly  improve DAP's performance. The attribute ``yellow''  is too trivial to detect with nearly 100\% accuracy already by DAP. The attribute ``swim'' is actually shared by visually similar categories, leaving not much room for KDICA to play any role.

\subsection{Zero-shot learning}
\label{sec:5.3}
As the intermediate representations of images and videos, attributes are often used in high-level computer vision applications. In this section, we conduct experiments on zero-shot learning to examine whether the improved attribute detectors could also benefit this task.

Given our UDICA and KDICA based attribute detection results, we simply input them to the second layer of the DAP model~\cite{lampert2009learning} to solve the  zero-shot learning problem. We then compare with several well-known zero-shot recognition systems as shown in Table~\ref{tab:zero}. We run our own experiments for some of them whose source code are provided by the authors. The corresponding results are again marked by \lq{}*\rq{}.

We see that in Table~\ref{tab:zero} the proposed simple solution to zero-shot learning outperforms the other state-of-the-art methods on the AWA, CUB, and UCF101 datasets, especially its immediate rival DAP. In addition, we notice that our kernel alignment technique (KDICA)  improves  the zero-shot recognition results over UDICA significantly on AWA. The improvements over UDICA on the other two datasets are also more significant than the improvements for the attribute prediction task (see Section~\ref{sec:5.2} and Table~\ref{tab:attribute}). This observation is interesting; it seems like implying that increasing the quality of the attribute detectors is rewarding, because the increase will be magnified to even larger improvement for the zero-shot learning.  Similar observation applies if we compare the differences between DAP and UDICA/KDICA respectively in Table~\ref{tab:zero} and Table~\ref{tab:attribute}. Finally, we note that our main purpose is indeed to investigate how better attribute detectors can benefit zero-shot learning. We do not expect to have a thorough comparison of the existing zero-shot learning methods.



\subsection{Multi-attribute based image retrieval}
\label{sec:5.4}

In this section, we do some experiments on the AWA  dataset for the multi-attribute based image retrieval, whose performance depends on the reliabilities of the attribute predictions. We input our learned feature representations to two popular frameworks for multi-attribute based image retrieval: TagProp~\cite{guillaumin2009tagprop} and the fusion of individual prediction scores~\cite{kumar2009attribute}. In TagProp, we use its $\sigma$ML variant to compute the ranking scores  of the multi-attributes queries. For the fusion of individual classifiers, we directly sum up the confidence scores corresponding to the multiple attributes in a query. The results of the fusion and TagProp are respectively shown in Table~\ref{tab:indi} and Table~\ref{tab:Tagpro}. We can  observe that our attribute-oriented representations improve the fusion technique for image retrieval on a variety  of queries (single attribute, attribute pairs, and triplets). Under the TagProp framework, the improvement is marginal on querying by attribute pairs and triples and  significant for single-attribute queries. 


\begin{table}
    \centering

    \begin{tabular}{c|ccc}

   \hline
   \hline
      query & VGG & UDICA    & KDICA \\

    \hline

        single  & 78.9 & 83.9 & \textbf{84.4} \\

        double  & 77.2 & 79.5 & \textbf{81.0} \\

        triple  & 76.1 & 78.6 & \textbf{79.4} \\

   \hline
   \hline
\end{tabular}
\vspace{3mm}
 \caption{Multi-attribute based image retrieval results on AWA by the late fusion of individual attribute detection scores. (\%, in AUC)}
\label{tab:indi}

\end{table}

\begin{table}
    \centering

    \begin{tabular}{c|ccc}
   \hline
   \hline
      query & VGG & UDICA    & KDICA \\
   \hline

        single  & 76.3 & 78.5 & \textbf{79.2} \\

        double  & 75.9 & 76.1 &  76.1 \\

        triple  & 75.5 & 75.6 & \textbf{75.8}\\

\hline
\hline
\end{tabular}
\vspace{3mm}
 \caption{Multi-attribute based image retrieval results on AWA by TagProp. (\%, in AUC)}
\label{tab:Tagpro}

\end{table}

\section{Conclusion}
\label{sec:6}
In this paper, we propose to re-examine the fundamental attribute detection problem and develop a novel attribute-oriented feature representation by casting the problem as multi-source domain generalization, such that one can conveniently apply off-shelf classifiers to obtain high-quality attribute detectors. The attribute detectors learned from our new representation are capable of breaking the boundaries of object categories and generalizing well to unseen classes. Extensive experiment on four datasets, and three tasks, validate that our attribute representation not only improves the quality of attributes, but also benefits succeeding applications, such as zero-shot recognition and image retrieval.

\pheadB{Acknowledgement}
This work was supported in part by NSF IIS-1566511. Chuang Gan was partially supported by the National
Basic Research Program of China Grant 2011CBA00300, 2011CBA00301, the National Natural Science Foundation of China Grant 61033001, 61361136003.
Tianbao Yang was partially supported by NSF IIS-1463988 and NSF IIS-1545995.

\end{document}